\documentclass[conference]{IEEEtran}
\IEEEoverridecommandlockouts
\usepackage{amsmath,amssymb,amsfonts}
\usepackage{algorithmic}
\usepackage{graphicx}
\usepackage{textcomp}
\usepackage{booktabs}
\usepackage{xcolor}
\usepackage{cite}
\usepackage{float}
\usepackage{subcaption}

\def\BibTeX{{\rm B\kern-.05em{\sc i\kern-.025em b}\kern-.08em
    T\kern-.1667em\lower.7ex\hbox{E}\kern-.125emX}}
    
\usepackage{lineno}
\begin{document}

\title{Unsupervised Anomaly Detection in Multi-Agent Trajectory Prediction via Transformer-Based Models}

\author{%
\IEEEauthorblockN{\fontsize{9}{10} Qing Lyu}
\IEEEauthorblockA{\fontsize{8}{9}\selectfont
\textit{Department of}\\
\textit{Civil and Environmental Engineering}\\
\textit{University of California, Berkeley}\\
Berkeley, USA \\
qinglyu@berkeley.edu}
\and
\IEEEauthorblockN{\fontsize{9}{10} Zhe Fu$^{*}$}
\IEEEauthorblockA{\fontsize{8}{9}\selectfont
\textit{Department of}\\
\textit{Civil and Environmental Engineering}\\
\textit{Electrical Engineering and Computer Sciences}\\
\textit{University of California, Berkeley}\\
Berkeley, USA \\
zhefu@berkeley.edu}
\and
\IEEEauthorblockN{\fontsize{9}{10} Alexandre Bayen}
\IEEEauthorblockA{\fontsize{8}{9}\selectfont
\textit{Department of}\\
\textit{Electrical Engineering and Computer Sciences}\\
\textit{Civil and Environmental Engineering}\\
\textit{University of California, Berkeley}\\
Berkeley, USA \\
bayen@berkeley.edu}
\thanks{*Corresponding author. Email: zhefu@berkeley.edu}
}

\maketitle

\begin{abstract}
Identifying safety-critical scenarios is essential for autonomous driving, but the rarity of such events makes supervised labeling impractical. Traditional rule-based metrics like Time-to-Collision are too simplistic to capture complex interaction risks, and existing methods lack a systematic way to verify whether statistical anomalies truly reflect physical danger.
To address this gap, we propose an unsupervised anomaly detection framework based on a multi-agent Transformer that models normal driving and measures deviations through prediction residuals. 
A dual evaluation scheme has been proposed to assess both detection stability and physical alignment: Stability is measured using standard ranking metrics in which Kendall Rank Correlation Coefficient captures rank agreement and Jaccard index captures the consistency of the top-$K$ selected items; Physical alignment is assessed through correlations with established Surrogate Safety Measures (SSM).
Experiments on the NGSIM dataset demonstrate our framework’s effectiveness: We show that the maximum residual aggregator achieves the highest physical alignment while maintaining stability. Furthermore,  our framework identifies 388 unique anomalies missed by Time-to-Collision and statistical baselines, capturing subtle multi-agent risks like reactive braking under lateral drift. The detected anomalies are further clustered into four interpretable risk types, offering actionable insights for simulation and testing.
\end{abstract}


\section{Introduction}

Ensuring safety in autonomous driving requires identifying \textit{safety-critical scenarios}, which are rare but high-impact events in naturalistic driving data. The scarcity of such events, known as the \textit{curse of rarity}~\cite{liu2024curse}, makes supervised labeling infeasible and motivates unsupervised methods that automatically detect risky behaviors from trajectories.

Traditional rule-based safety indicators, such as Time-to-Collision (TTC) and Deceleration Rate to Avoid Crash (DRAC)~\cite{fu2021comparison}, are simple and interpretable but limited to pairwise interactions. Data-driven methods improve flexibility but often rely on single-agent reconstruction or heuristic thresholds, lacking robust scene-level reasoning.

To address these limitations, we propose a Transformer-based unsupervised anomaly detection framework that explicitly models multi-agent interactions and evaluates both statistical consistency and alignment with interpretable physical risk indicators. The main contributions are:
\begin{itemize}
    \item An unsupervised detection pipeline that integrates a multi-agent Transformer with a robust residual aggregation and Isolation Forest scoring method to identify statistically anomalous scenarios.
    \item A novel dual evaluation framework that ensures stable and physically meaningful detection of safety-critical scenes.
\end{itemize}

This framework provides a unified, label-free process for discovering safety-critical driving scenarios in large-scale naturalistic datasets.

\section{Related Work}

\subsection{Safety-Critical Scenario Detection}

A key challenge in autonomous vehicle (AV) validation is identifying rare but safety-critical ``corner" or ``edge" cases that disproportionately impact overall risk and testing cost~\cite{liu2024curse}. Naturalistic driving data are highly imbalanced, making empirical safety evaluation statistically unreliable. Thus, systematic methods are needed to automatically surface these high-impact scenarios with high recall and minimal labeling, enabling targeted testing and model improvement.

\paragraph{Classical Approaches}
Early scenario detection relied on \textit{rule-based surrogate safety measures} (SSMs) that estimate collision risk from physical indicators such as Time-to-Collision (TTC), Deceleration Rate to Avoid Crash (DRAC), and Proportion of Stopping Distance (PSD)~\cite{gettman2003surrogate}. 
Threshold rules remain popular for their simplicity and interpretability, but are restricted to pairwise interactions and prone to false positives in complex, multi-agent traffic. 
As driving environments become increasingly heterogeneous, these fixed heuristics lack generality and fail to anticipate unseen unsafe behaviors~\cite{lefevre2014survey}.

\paragraph{Data-Driven Approaches}
To address the limits of heuristic metrics, recent work adopts \textit{data-driven} methods that learn risk patterns directly from data~\cite{devika2024vadgan}. 
Two main paradigms exist: (1) \textit{scenario generation}, which uses generative or reinforcement learning (RL) models to synthesize rare but plausible high-risk events, and (2) \textit{scenario detection}, which mines naturalistic datasets for existing critical cases~\cite{kibalama2022av}. 
RL-based methods train adversarial agents to provoke AV failures: such as forced cut-ins or abrupt braking, thereby exposing system weaknesses~\cite{liu2024curse}. 
Meanwhile, scenario detection targets large unlabeled datasets, where safety-critical events are too rare to label manually. 
This motivates unsupervised frameworks that model normal driving behavior and use prediction residuals as a deviation signal for anomaly scoring.

\subsection{Trajectory Prediction}

Trajectory prediction is fundamental to autonomous driving, enabling vehicles to anticipate surrounding agents for safe planning and control. Existing methods can be broadly classified into \textit{physics-based}, \textit{machine learning-based}, and \textit{deep learning-based} approaches~\cite{huang2022survey}.

\paragraph{Physics-based and Classical Machine Learning Models.}
Early work used kinematic and dynamic models such as Constant Velocity or Constant Turn Rate~\cite{ammoun2009real}. Later methods applied Kalman Filters, Gaussian processes, and Monte Carlo–MPC frameworks to handle uncertainty~\cite{prevost2007extended}. Statistical approaches with GPs, HMMs, and SVMs modeled driver behavior but relied on predefined motion categories, limiting performance in dense traffic~\cite{guo2019modeling}.

\paragraph{Deep and Transformer-based Models}
Deep models dominate trajectory prediction. RNNs and LSTMs capture temporal dependencies~\cite{graves2013generating}, CNNs extract local spatiotemporal patterns~\cite{nikhil2018convolutional}, and GNNs model agent–map interactions~\cite{wu2020comprehensive}. Generative approaches like Social-GAN~\cite{goodfellow2014generative} handle multimodal outcomes. Transformers further improve by modeling long-range and multi-agent dependencies~\cite{vaswani2017attention}. Recent frameworks such as Trajectron++~\cite{salzmann2020trajectron++}, SceneTransformer~\cite{ngiam2021scene}, and MultiPath++~\cite{varadarajan2022multipath++} integrate map semantics and multimodal decoding for diverse, uncertain future predictions.

\subsection{Anomaly Detection in Trajectories}

Anomaly detection identifies rare behaviors that deviate from normal driving patterns, serving as a key unsupervised tool for discovering edge cases and safety risks~\cite{wang2021machine}.

\paragraph{Traditional Methods}
Early methods relied on handcrafted distance or density metrics, such as Dynamic Time Warping (DTW)~\cite{langfu2023method}, edit distance~\cite{burnaev2016conformalized}, and density- or clustering-based models like TRAOD~\cite{lee2007trajectory} and DBSCAN~\cite{zhang2021ut}. These approaches are interpretable but sensitive to feature design and often fail with noisy or high-dimensional data.

\paragraph{Reconstruction Based Methods}
Reconstruction based models, typically Autoencoders (AEs), learn to reproduce normal trajectories and use \textit{reconstruction error} as the anomaly score. Variants such as LSTM AEs~\cite{di2018unsupervised} and CNN--LSTM hybrids~\cite{kieu2018outlier} capture temporal features automatically. Fan et al.~\cite{fan2022hybrid} proposed a hybrid framework based on recurrent--convolutional autoencoders and one-class classifiers to identify anomalous lane-changing behaviors from naturalistic driving data. These models are usually trained on normal data and tend to yield larger reconstruction errors on abnormal behaviors.

\paragraph{Prediction Error Based Methods}
Prediction error-based methods identify anomalies by comparing predicted and actual data values. The basic idea is that normal patterns can be accurately predicted, while abnormal events cause large prediction errors. Early approaches used models such as LSTMs to forecast future values and flagged large deviations as anomalies. Recent studies have improved this idea by combining prediction with reconstruction to increase robustness ~\cite{tang2020integrating}, learning false positive patterns to refine anomaly scores~\cite{li2020anomaly}, and using multi-frame prediction errors to better distinguish true anomalies in videos~\cite{kim2025mpe}.

\subsection{Research Gap}

Existing methods face two key limitations. First, most rely on rule-based surrogate measures or single-agent prediction models, making them unable to capture the multi-agent interactions that produce many safety-critical events. Consequently, their anomaly scores remain local or heuristic rather than truly scene-level.
Second, current prediction-error approaches seldom verify whether detected anomalies correspond to real physical danger. They lack a systematic way to determine whether large residuals align with safety indicators such as low Time-to-Collision or small spacing, leaving the connection between statistical abnormality and physical risk largely unvalidated.

To overcome these limitations, we propose an unsupervised anomaly detection framework that (i) uses a multi-agent Transformer with structured residual aggregation to produce coherent scene-level anomaly scores, and (ii) introduces a dual evaluation pipeline that jointly assesses anomaly stability and physical-risk alignment. This yields a principled and physically grounded approach for discovering safety-critical scenes at scale.

\section{Methodology}
\begin{figure}[htbp]
    \centering
    \includegraphics[width=0.48\textwidth]{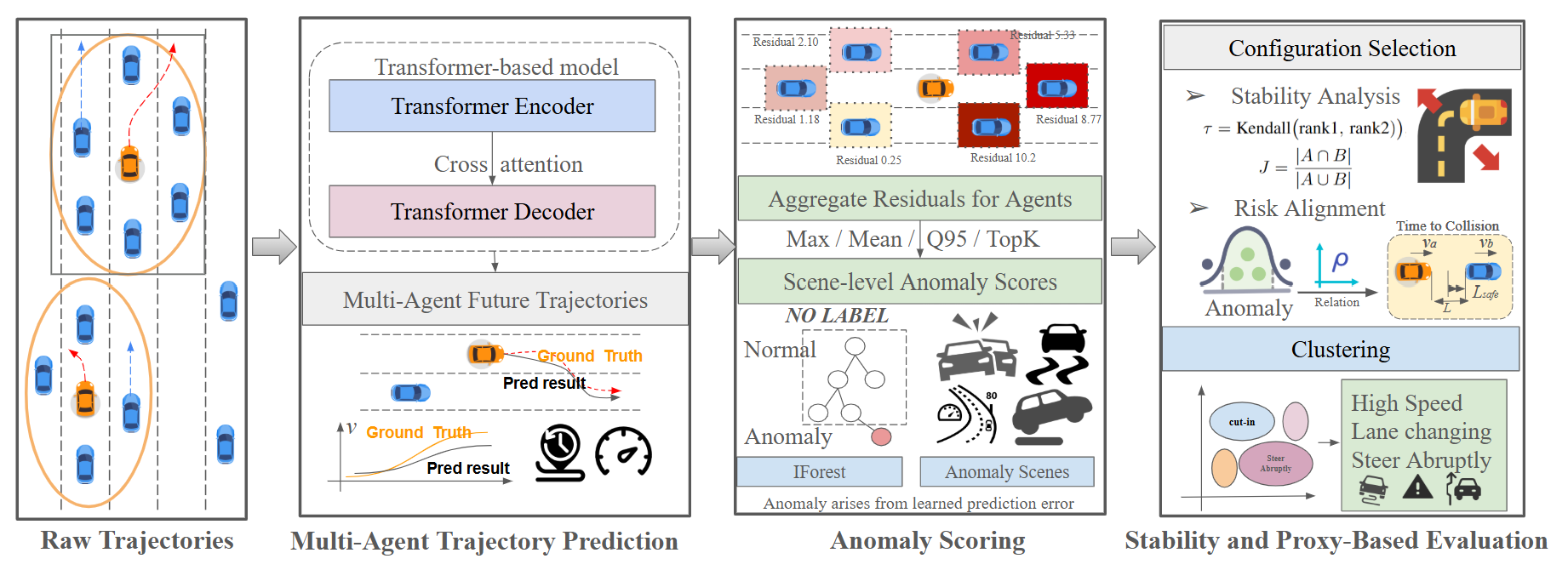}
    \caption{Overview of unsupervised anomaly detection pipeline.}
    \label{fig:pipeline}
\end{figure}

To provide an overview, the entire unsupervised anomaly detection pipeline is illustrated in Fig.~\ref{fig:pipeline}.

\subsection{Predictive Modeling}

The predictive model forms the basis of our anomaly detection framework, using prediction errors to quantify behavioral deviations from normal driving patterns.

% \subsubsection{Problem Setup}
% We formulate multi-agent trajectory forecasting as a sequence prediction task.
% For each scene with $A$ vehicles, the model observes $T_{\text{enc}}$ past steps and predicts the next $T_{\text{pred}}$ steps.
% During training, the decoder receives $T_{\text{label}}$ future steps as context to stabilize prediction.

\subsubsection{Model Architecture}
\begin{figure}[htbp]
    \centering
    \includegraphics[width=\linewidth]{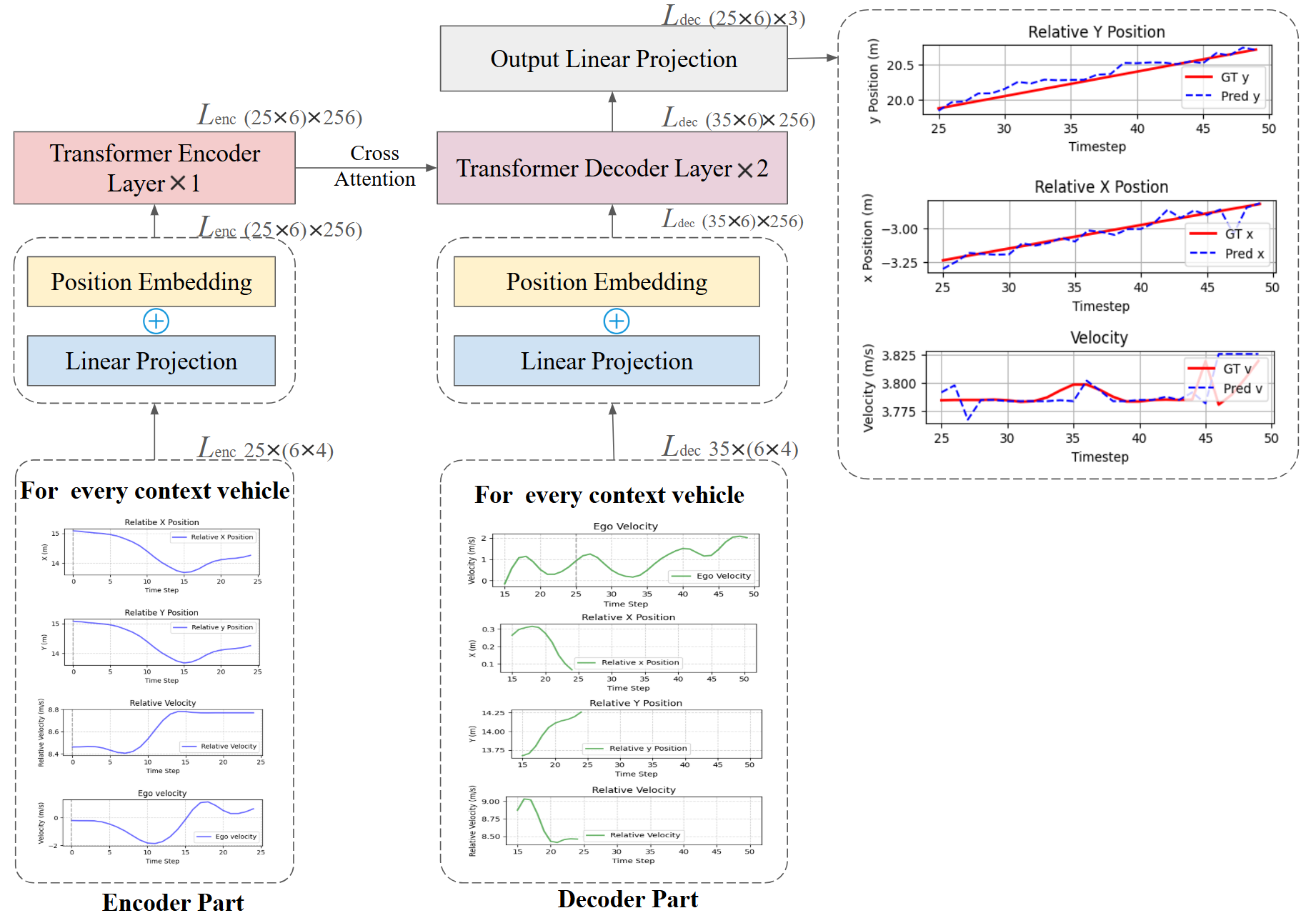}
    \caption{Structure of the Transformer-based trajectory predictor.}
    \label{fig:transformer_architecture}
\end{figure}

We adopt a sequence-to-sequence Transformer architecture for multi-agent motion prediction~\cite{zhu2022transfollower}. 
The model encodes $T_{\text{enc}}$ historical states of all interacting vehicles and decodes future trajectories over a horizon of $T_{\text{pred}}$ steps. 
Following an encoder–decoder structure, it learns past multi-agent motion patterns to future relative displacements and velocities, as illustrated in Fig.~\ref{fig:transformer_architecture}. This study focuses on the overall detection pipeline for identifying safety-critical anomalous scenes.

\paragraph{Encoder}
The encoder processes past multi-agent motion to capture temporal and spatial dependencies. 
Each historical state $\mathbf{X}_{\text{enc}} \in \mathbb{R}^{H\times 3}$ is linearly projected to a 256-dimensional embedding and passed through two Transformer encoder layers, each consisting of multi-head self-attention and feed-forward blocks. 
The resulting representation $\mathbf{Z}_{\text{enc}} \in \mathbb{R}^{H\times 256}$ summarizes the interaction-aware historical context for the decoder.

\paragraph{Decoder}

The decoder generates future trajectories conditioned on the encoded past.
Its input includes the last $T_{\text{label}}$ observed frames and $T_{\text{pred}}$ placeholder future frames, each embedded with learned positional encodings.
Self-attention captures temporal structure within the predicted horizon, while cross-attention injects interaction-aware context from $\mathbf{Z}_{\text{enc}}$. 
A final projection layer outputs the predicted coordinates $(\Delta x, \Delta y, v)$ for all agents over future time steps.

\subsubsection{Training Objective}
Trajectory features $(\Delta x, \Delta y, v)$ are standardized using training-set statistics. 
The model is trained with a weighted mean squared error over future positions and velocities:
\[
\mathcal{L} =
\lambda_{\text{pos}}\|\hat{\mathbf{p}}-\mathbf{p}\|_2^2 +
\lambda_{\text{vel}}\|\hat{\mathbf{v}}-\mathbf{v}\|_2^2,
\]
where $\lambda_{\text{pos}}$ and $\lambda_{\text{vel}}$ balance spatial and kinematic accuracy, producing stable predictions for downstream anomaly scoring. We set $\lambda_{\text{pos}} = 1.0$ and $\lambda_{\text{vel}} = 0.5$ in all experiments to balance spatial and kinematic accuracy.

\subsection{Prediction-Error–Driven Anomaly Scoring}
For each agent $a$ at time $t$, the predictor outputs a forecasted position 
$\hat{\mathbf{p}}_{a,t}$ and velocity $\hat{v}_{a,t}$, which are compared with ground truth to compute instantaneous deviations:
\[
e_{a,t}^{\text{pos}} = \|\hat{\mathbf{p}}_{a,t} - \mathbf{p}_{a,t}\|_2, 
\qquad
e_{a,t}^{\text{vel}} = |\hat{v}_{a,t} - v_{a,t}|.
\]
A scalar residual is defined as a weighted combination of spatial and kinematic errors:
\[
e_{a,t} =
\alpha_{\text{pos}}\, e_{a,t}^{\text{pos}} +
\alpha_{\text{vel}}\, e_{a,t}^{\text{vel}}.
\]
Here, $\alpha_{\text{pos}}$ and $\alpha_{\text{vel}}$ weight the spatial and
kinematic error terms. 
We set $\alpha_{\text{pos}} = 1.0$ and $\alpha_{\text{vel}} = 0.5$ in all experiments.

To obtain scene-level anomaly intensity, we aggregate residuals $\{e_{a,t}\}$ over all agents and timesteps:
\[
S_{\text{scene}}^{(f)} = f\!\left(\{e_{a,t}\}_{a,t}\right), \qquad
f \in \{\text{max},\, \text{q95},\, \text{mean},\, \text{top-}k\}.
\]

Here, \textit{max} returns the largest residual; \textit{q95} computes the 95th percentile of the set; \textit{mean} is the average; and \textit{top-}$k$ computes the mean of the $k$ largest residuals. These alternatives are later compared to identify the most stable and physically aligned aggregation method.

\subsection{Unsupervised Anomaly Identification}

For each aggregation operator 
$f \in \{\text{max}, \text{q95}, \text{mean}, \text{top-}k\}$,
the scene-level scores $S_{\text{scene}}^{(f)}$ are fed into an unsupervised Isolation Forest to identify anomalous driving scenes.
We vary the contamination parameter over $\{0.10, 0.15, 0.20\}$ to examine different anomaly coverage levels. 
The contamination parameter is a key hyperparameter of Isolation Forest, as it specifies the expected proportion of outliers and thus determines the decision threshold for labeling samples as anomalous.

To assess the stability of each aggregation operator, we measure the
agreement between anomaly rankings produced under different contamination
levels. Kendall’s~$\tau$ quantifies rank correlation, while Jaccard@K
evaluates overlap among the top-$K$ detected scenes:
\[
\tau^{(f)} =
\text{Kendall }\!\big(\text{rank}(S^{(f,c_1)}),\, \text{rank}(S^{(f,c_2)})\big),
\]
\[
J^{(f)} =
\frac{|A^{(f,c_1)} \cap A^{(f,c_2)}|}
{|A^{(f,c_1)} \cup A^{(f,c_2)}|}.
\]
High values of $\tau^{(f)}$ and $J^{(f)}$  indicate consistent anomaly
rankings across operating points, reflecting a stable and robust
unsupervised detection method.

\subsection{Proxy-Aligned Metric Selection}

To assess whether detected anomalies correspond to physically hazardous events, we evaluate their alignment with interpretable surrogate safety measures, including harsh closing ratio, lateral excursion, minimum longitudinal gap, minimum time-to-collision, and relative speed variability. Residual-based anomalies reflect deviations from learned normal behavior rather than definitive risk labels; proxy alignment is used to quantify their physical consistency.

For each aggregation configuration, we compute Spearman’s rank correlation between 
the anomaly scores and each safety proxy:
\[
\rho = \text{corr}
\!\left(\text{proxy},\, \text{rank}(S_{\text{scene}})\right).
\]
High values of $\rho$ indicate that scenes assigned larger anomaly scores 
indeed exhibit greater physical risk.

Among all aggregation operators and contamination levels, we select the configuration with relatively higher and more consistent correlations to surrogate safety proxies while preserving stable rankings. This yields an anomaly score that is both robust and physically meaningful.

\subsection{Scenario Clustering and Interpretation}
For interpretability, anomalous scenes are clustered using K-Means on a 6D feature vector of lateral and longitudinal prediction error statistics. The number of clusters $K$ is selected by maximizing the Silhouette coefficient. The resulting clusters correspond to distinct behavioral patterns such as unstable following, sudden braking, and abrupt lane changes, providing structured interpretation of detected anomalies.

\section{Experimental Results}

\subsection{Dataset and Preprocessing}

\begin{figure}[htbp]
    \centering
    \begin{subfigure}[b]{0.45\linewidth}
        \centering
        \includegraphics[height=4cm, width=\linewidth]{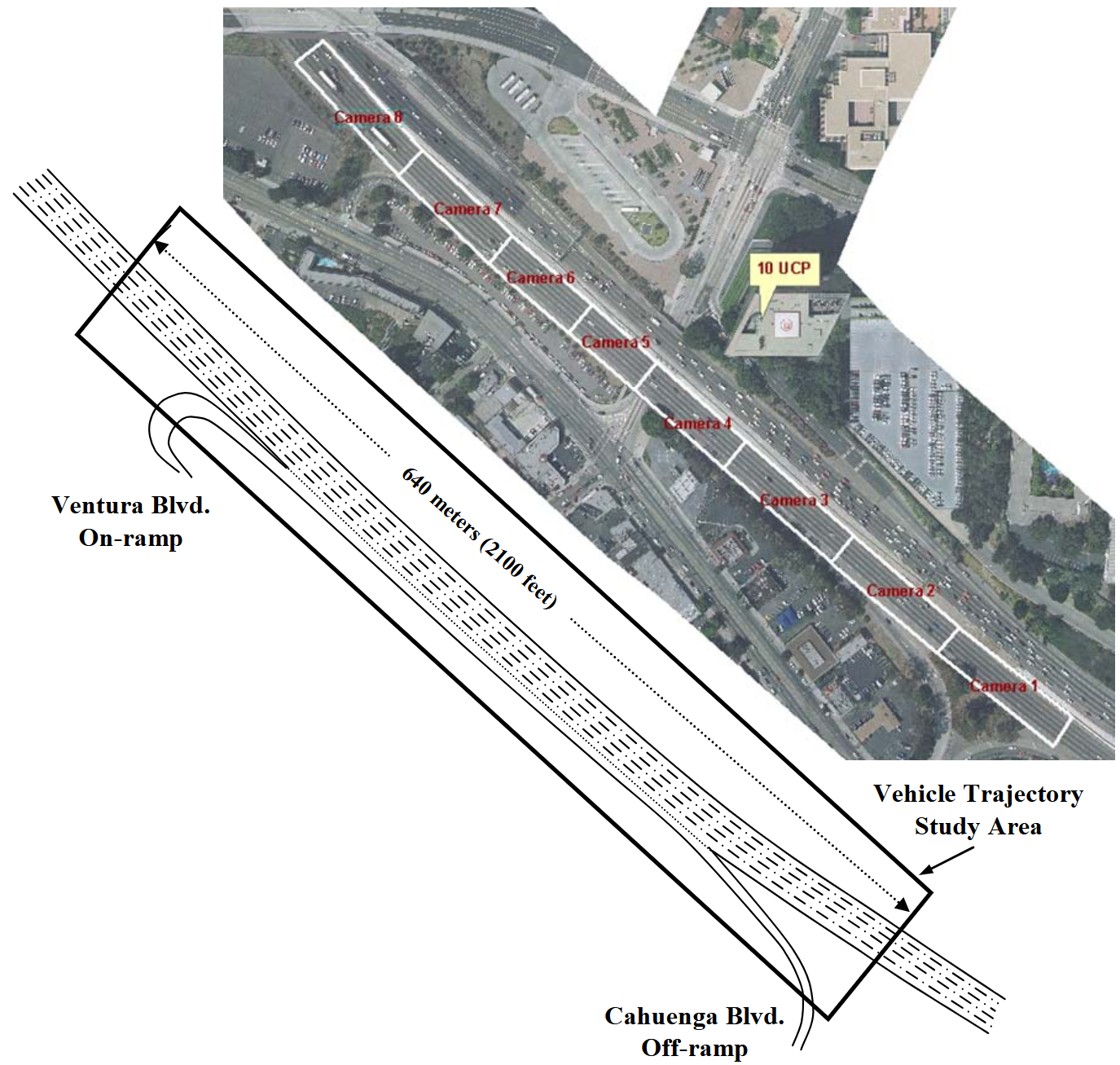}
        \caption{NGSIM US101 study area.}
        \label{fig:dataset_area}
    \end{subfigure}
    \hfill
    \begin{subfigure}[b]{0.5\linewidth}
        \centering
        \includegraphics[height=4cm,width=1.5cm]{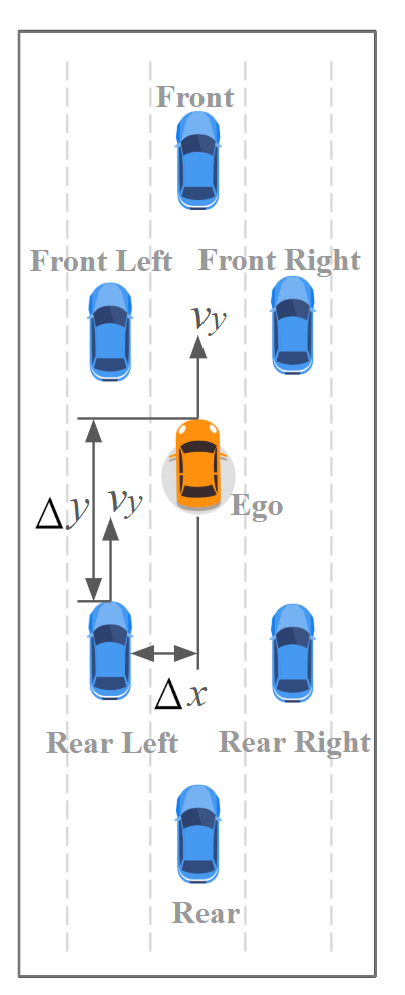}
        \caption{Ego-centric scene.}
        \label{fig:dataset_scene}
    \end{subfigure}
    \caption{Dataset overview and scene construction: 
    (a) the 640~m US101 study area, 
    (b) the ego-centric scene representation with one ego vehicle and six surrounding vehicles.}
    \label{fig:dataset}
\end{figure}

We use the NGSIM US101 dataset~\cite{coifman2017critical}, which provides 10~Hz vehicle trajectories over a 640~m freeway segment, as shown in Fig.~\ref{fig:dataset}. 
Each record contains position, velocity, acceleration, and lane identifiers for all vehicles. 
From these data, we extract 5-s ego-centric scenes including the ego and up to six surrounding vehicles (\textit{Front}, \textit{Rear}, \textit{Front-Left}, \textit{Front-Right}, \textit{Rear-Left}, \textit{Rear-Right}), as illustrated in Fig.~\ref{fig:dataset}(b). 
This configuration captures the most relevant local interactions for trajectory prediction and anomaly analysis.

We filter out noisy or degenerate samples: a scene is discarded if any agent shows a spatial jump over 10~m between frames or remains stationary throughout the window.

Each processed scene is a tensor of shape $50\times7\times3$ (time $\times$ agents $\times$ features), where the features are $(x, y, v)$ in the ego frame. 
The model consumes $T_{\text{enc}}=25$ past steps and predicts $T_{\text{pred}}=25$ future steps, with steps 15-25 used as decoder labels during training. 
See Fig.~\ref{fig:transformer_architecture} for the input dimensionality.

The final dataset includes 28{,}982 multi-agent scenes from 6{,}101 raw trajectories, split into 70\% training, 20\% validation, and 10\% testing. 
All trajectories are normalized relative to the ego frame before model training and inference.

\subsection{Prediction Performance}

We evaluate trajectory forecasting accuracy using \textit{Average Displacement Error} (ADE) and \textit{Final Displacement Error} (FDE):
{\small
\[
\text{ADE} = \frac{1}{NTA}\!\sum_{i,t,a}\|\hat{\mathbf{p}}_{a,t}^{(i)}-\mathbf{p}_{a,t}^{(i)}\|_2, \quad
\text{FDE} = \frac{1}{NA}\!\sum_{i,a}\|\hat{\mathbf{p}}_{a,T}^{(i)}-\mathbf{p}_{a,T}^{(i)}\|_2.
\]
}

We compare our Transformer model with a standard LSTM encoder--decoder baseline. We adopt the Transformer for its ability to capture long-range dependencies and multi-agent interactions~\cite{vaswani2017attention,zhu2022transfollower}, and observe lower prediction error than the LSTM baseline. Accordingly, the predictor is primarily used to generate reliable residual signals for downstream detection, while this work primarily focuses on the anomaly detection and evaluation framework.

All models are trained using the AdamW optimizer with an initial learning rate of $1\times10^{-3}$, decayed by a StepLR scheduler with step size 20 and decay factor $\gamma = 0.1$. 
Training uses a batch size of 32 with early stopping based on validation loss.

\begin{table}[htbp]
\caption{Prediction Performance Comparison}
\centering
\begin{tabular}{lcc}
\toprule
\textbf{Model} & \textbf{ADE (m) ↓} & \textbf{FDE (m) ↓} \\
\midrule
Transformer-based (ours) & 1.28 & 1.59 \\
LSTM Encoder--Decoder & 1.95 & 2.19 \\
\bottomrule
\end{tabular}
\label{tab:pred_perf}
\end{table}

\begin{figure}[htbp]
    \centering
    \begin{subfigure}[b]{0.48\linewidth}
        \centering
        \includegraphics[width=\linewidth]{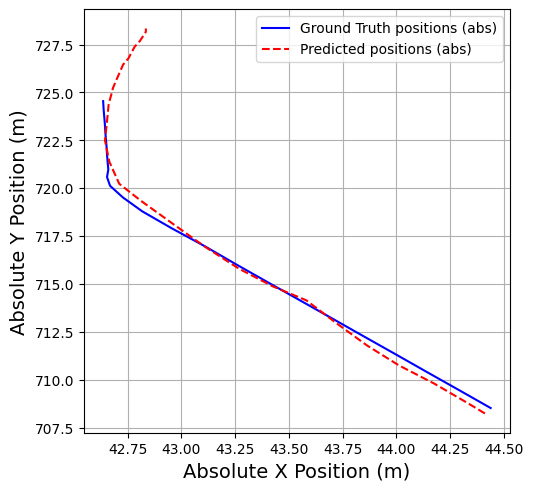}
        \caption{Trajectory comparison between ground-truth and predicted positions.}
        \label{fig:traj_comp}
    \end{subfigure}
    \hfill
    \begin{subfigure}[b]{0.46\linewidth}
        \centering
        \includegraphics[width=\linewidth]{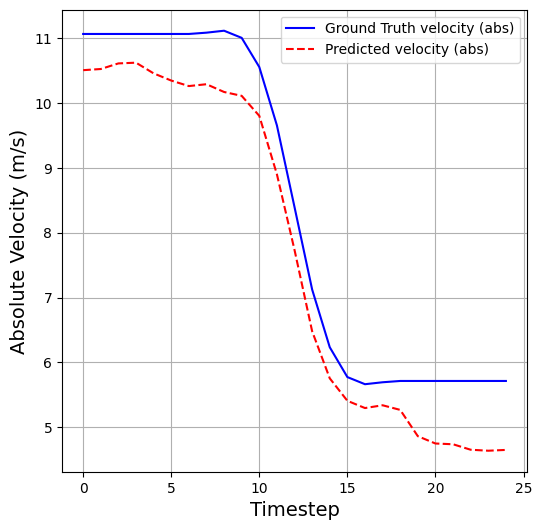}
        \caption{Velocity comparison between ground-truth and predicted profiles.}
        \label{fig:vel_comp}
    \end{subfigure}
    \caption{Example qualitative results of the Transformer-based trajectory predictor}
    \label{fig:prediction_examples}
\end{figure}

Qualitative examples of trajectory and velocity predictions are shown in Fig.~\ref{fig:prediction_examples}. 
The predicted trajectories closely follow the ground truth, demonstrating the model’s ability to capture both spatial and temporal motion patterns.

\subsection{Scene-Level Anomaly Detection}

\begin{figure}[htbp]
    \centering
    \includegraphics[width=0.55\linewidth]{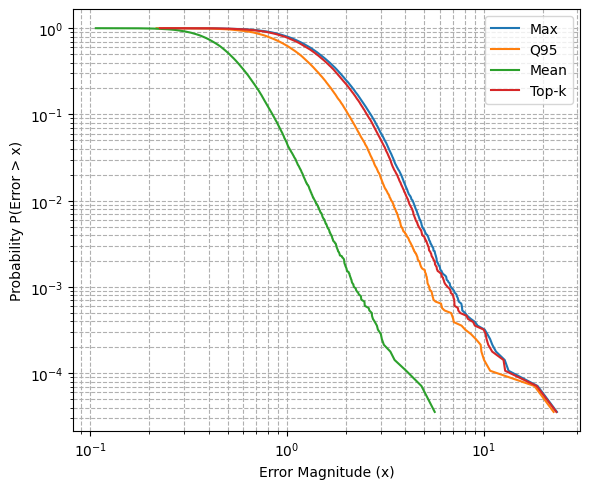}
    \caption{CCDF of scene-level anomaly scores on a log--log scale.}
    \label{fig:ccdf}
\end{figure}

Scene-level anomaly scores, obtained by aggregating per-agent residuals (\textit{max}, \textit{mean}, \textit{q95}, and \textit{top-$k$}), exhibit a clear long-tailed distribution, as illustrated in Fig.~\ref{fig:ccdf}. Most scenes follow consistent dynamics, while a few deviate sharply from normal behavior. This heavy-tail pattern justifies applying an Isolation Forest to capture rare, high-deviation events.

Detection consistency across contamination levels is evaluated using 
Kendall’s~$\tau$ and Jaccard@K. 
As summarized in Table~\ref{tab:stability}, all aggregation operators show 
high agreement ($\tau>0.95$, Jaccard@K$ >0.95$), indicating that varying 
the cutoff proportion affects coverage but not ranking, 
and the same high-risk scenes are consistently identified.

\begin{table}[htbp]
\caption{Ranking consistency across contamination levels.}
\centering
\begin{tabular}{lcc}
\toprule
\textbf{Aggregator} & \textbf{Kendall $\tau$~($\uparrow$)} & 
\textbf{Jaccard@K~($\uparrow$)} \\
\midrule
mean   & 0.98 & 0.97  \\
q95    & 0.97 & 0.98 \\
top-$k$ & 0.98 & 0.98  \\
max    & 0.99 & 1.00 \\
\bottomrule
\end{tabular}
\label{tab:stability}
\end{table}

\subsection{Proxy-Aligned Correlation Analysis}

We evaluate how the learned anomaly scores correspond to physically interpretable safety proxies, including harsh closing ratio, lateral excursion, minimum longitudinal gap, minimum TTC, and relative speed variability. Spearman’s~$\rho$ is computed between the Isolation Forest scores and each proxy, and results are averaged across contamination levels ($c=0.10, 0.15, 0.20$).

% As summarized in Table~\ref{tab:proxy_corr}, the max and q95 aggregators achieve relatively higher correlations with several safety proxies than alternative aggregation strategies. These results highlight the limited ability of rule-based surrogate measures to capture complex, multi-agent risk patterns, supporting our motivation that handcrafted proxies are insufficient for realistic traffic scenes. 
% Higher anomaly scores are generally associated with smaller minimum TTC and longitudinal gaps, and larger relative speed variability and harsh closing ratios, indicating meaningful physical consistency. 
% Based on stability and proxy alignment, we adopt the max aggregator with $c=0.15$ for subsequent clustering, and qualitative examples further support that high-scoring scenes reflect safety-relevant interaction patterns.

Table~\ref{tab:proxy_corr} shows moderate proxy alignment, which is expected because rule-based SSMs are largely pairwise and cannot fully represent complex multi-agent or lateral risks. Higher anomaly scores are generally associated with smaller minimum TTC and longitudinal gaps, and larger relative speed variability and harsh closing ratios. Among aggregators, max and q95 yield the most consistent correlations across multiple proxies, so we adopt max with $c=0.15$ for downstream analysis.

\begin{table}[!t]

\caption{Average Spearman's $\rho$ between anomaly scores and safety proxies (averaged across contamination levels $c = 0.10, 0.15, 0.20$).}
\label{tab:proxy_corr}
\centering
\begin{tabular}{lcccc}
\toprule
\textbf{Safety Proxy} & \textbf{max} & \textbf{mean} & \textbf{q95} & \textbf{top-$k$} \\
\midrule
Harsh closing ratio & 0.312 & 0.088 & 0.290 & 0.256 \\
Lateral excursion (m) & 0.139 & 0.156 & 0.128 & 0.136 \\
Min longitudinal gap (m) & -0.162 & -0.120 & -0.137 &  -0.156 \\
Min TTC (s) &  -0.235 & -0.085 &  -0.108 &  -0.211 \\
Relative speed std (m/s) &  0.298 &  0.194 &  0.304 &  0.281 \\
\bottomrule
\end{tabular}
\end{table}

\subsection{Scenario Clustering and Ensemble Interpretation}
\begin{figure}[ht]
    \centering
    \begin{subfigure}[b]{0.38\columnwidth}
        \centering
        \includegraphics[height=2.8cm, keepaspectratio]{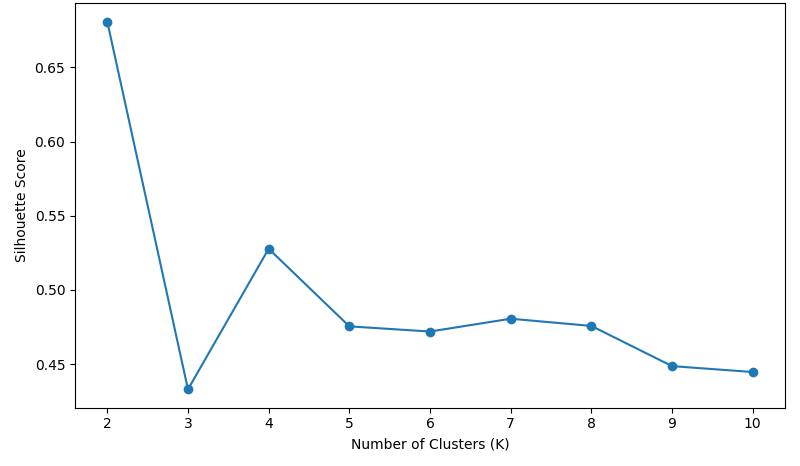}
        \caption{Silhouette scores}
        \label{fig:silhouette}
    \end{subfigure}
    \hfill
    \begin{subfigure}[b]{0.42\columnwidth}
        \centering
        \includegraphics[height=2.8cm, keepaspectratio]{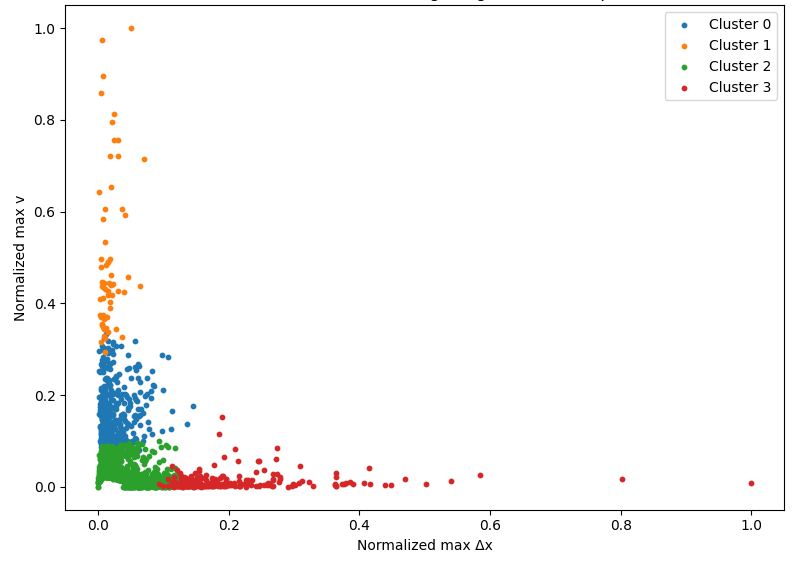}
        \caption{Clusters visualization}
        \label{fig:anomaly_cluster}
    \end{subfigure}
    \caption{KMeans clustering of anomaly scenes}
    \label{fig:cluster_analysis}
\end{figure}

The anomalous scenes selected by the max aggregator with $c=0.15$ are further clustered 
using K-Means on a compact 6D feature vector composed of lateral and longitudinal 
prediction errors. The optimal number of clusters, $K=4$, is determined by maximizing the Silhouette coefficient~\cite{rousseeuw_silhouettes_1987}, 
as shown in Fig.~\ref{fig:cluster_analysis}. 

% The clustering results reveal four distinct types of motion deviation:
% \begin{itemize}
%     \item C0: Unstable following with longitudinal oscillations %(Fig.~\ref{fig:cluster0_case}).
%     \item C1: Abrupt acceleration or braking %(Fig.~\ref{fig:cluster1_case}).
%     \item C2: Mild lateral drift with steady motion %(Fig.~\ref{fig:cluster2_case}).
%     \item C3: Sharp lane change or evasive maneuver %(Fig.~\ref{fig:cluster3_case}).
% \end{itemize}
% These clusters reflect different combinations of lateral and longitudinal error statistics.

We compare the physical risk metrics between normal and anomalous scenes, as summarized in Table~\ref{tab:anomaly_vs_normal}.
Anomalies show lower TTC and spacing but higher relative velocity and acceleration, indicating greater risk and instability.

\begin{table}[ht]
\centering
\caption{Comparison between Anomalous and Normal Scenes}
\label{tab:anomaly_vs_normal}
\begin{tabular}{lcc}
\toprule
\textbf{Metric} & \textbf{Anomalous} & \textbf{Normal} \\
\midrule
Min TTC (s) & $1.48 \pm 0.24$ & $2.56 \pm 0.32$ \\
Min Dist (m) & $12.78 \pm 3.38$ & $15.59 \pm 2.95$ \\
Max $\Delta v$ (m/s) & $23.3 \pm 14.6$ & $18.5 \pm 11.8$ \\
Max Acc (m/s$^2$) & $4.85 \pm 3.82$ & $2.93 \pm 2.33$ \\
\bottomrule
\end{tabular}
\end{table}

To better characterize the behavioral patterns underlying the detected anomalies, 
we examine the cluster centers in the min--max normalized 6D prediction-error feature space 
(max/mean/std of lateral position error $\Delta x$ and velocity error $v$), as shown in 
Table~\ref{tab:cluster_center_normalized}. These centers summarize representative deviation profiles 
and support the following interpretations:

\begin{table}[ht]
\centering
\caption{Cluster Centers of Prediction-Based Anomalous Scene (Min-Max Normalized)}
\label{tab:cluster_center_normalized}
\begin{tabular}{lcccccc}
\toprule
\textbf{Cluster} & \textbf{Max $\Delta x$} & \textbf{Mean $\Delta x$} & \textbf{Std $\Delta x$} & \textbf{Max $v$} & \textbf{Mean $v$} & \textbf{Std $v$} \\
\midrule
Cluster 0 & 0.024 & 0.026 & 0.025 & 0.159 & 0.175 & 0.157 \\
Cluster 1 & 0.018 & 0.022 & 0.018 & 0.500 & 0.531 & 0.496 \\
Cluster 2 & 0.048 & 0.042 & 0.049 & 0.022 & 0.029 & 0.021 \\
Cluster 3 & 0.194 & 0.152 & 0.201 & 0.011 & 0.016 & 0.010 \\
\bottomrule
\end{tabular}

\end{table}

\begin{itemize}
    \item \textbf{Cluster 0 (Car-following instability):} low lateral error but moderate velocity error statistics, consistent with oscillatory longitudinal pacing.
    \item \textbf{Cluster 1 (Sudden acceleration/braking):} the largest velocity-error magnitude and variance with minimal lateral error, indicating abrupt longitudinal maneuvers.
    \item \textbf{Cluster 2 (Mild lateral deviation):} elevated lateral error with low velocity variation, suggesting gradual drift or mild lateral offsets.
    \item \textbf{Cluster 3 (Abrupt lateral maneuver):} extreme lateral error with negligible velocity error, consistent with sharp lane changes or evasive swerves at roughly steady speed.
\end{itemize}

\paragraph{\textit{Case Study: Cluster 0.}}

\begin{figure}[htbp]
\centering
\begin{subfigure}[b]{0.15\textwidth}
\centering
\includegraphics[height=2.0cm,  width=\textwidth]{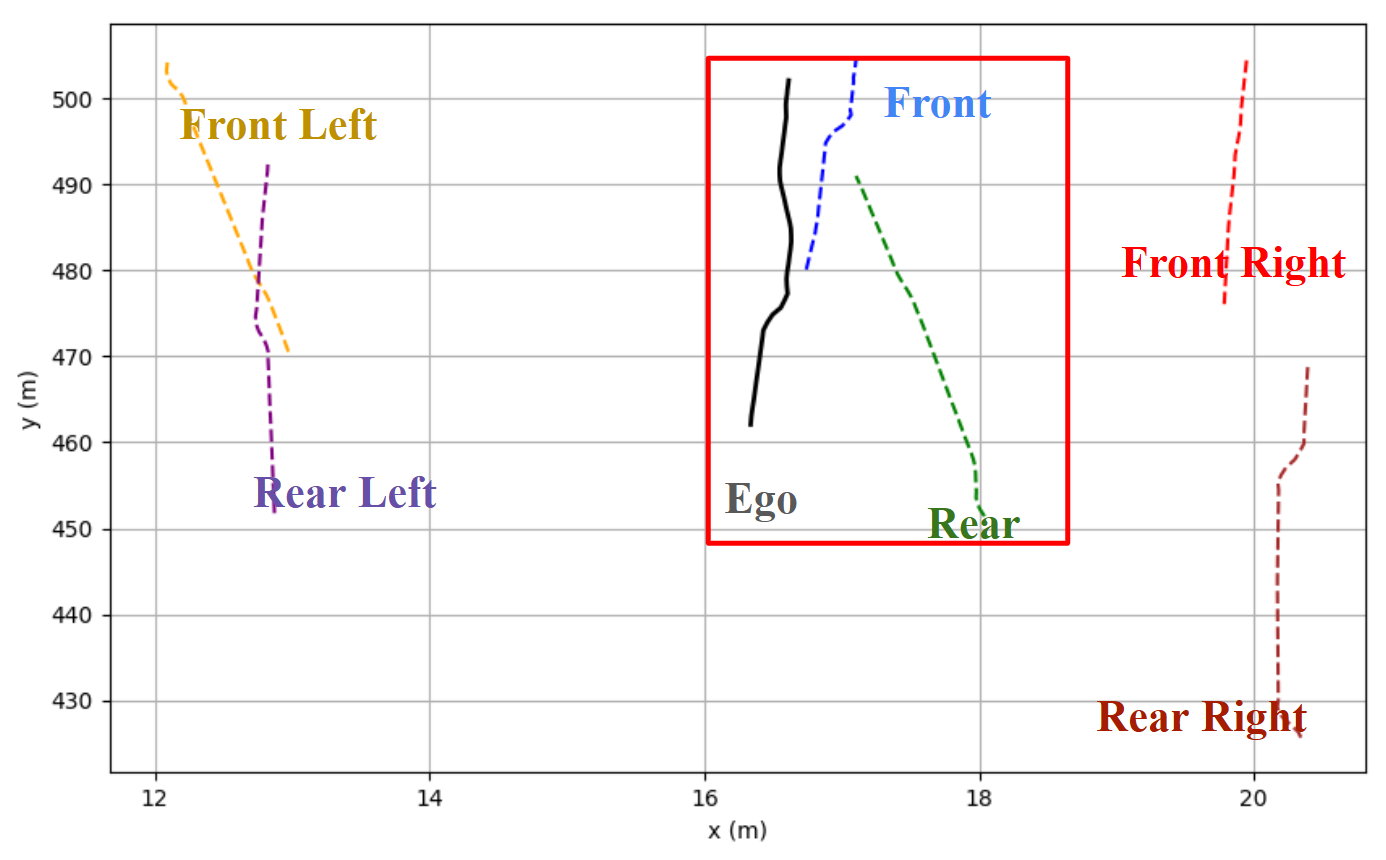}
\caption{Trajectory Overview}
\end{subfigure}
\hfill
\begin{subfigure}[b]{0.15\textwidth}
\centering
\includegraphics[height=2.0cm, width=\textwidth]{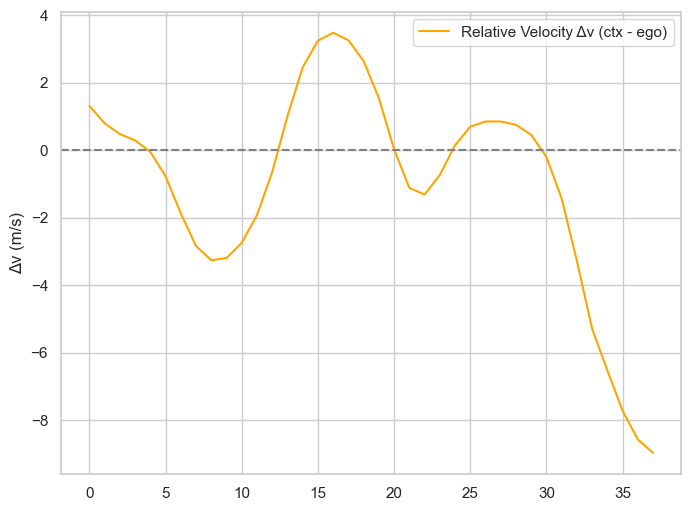}
\caption{Relative Velocity ($\Delta v$)}
\end{subfigure}
\hfill
\begin{subfigure}[b]{0.15\textwidth}
\centering
\includegraphics[height=2.0cm, width=\textwidth]{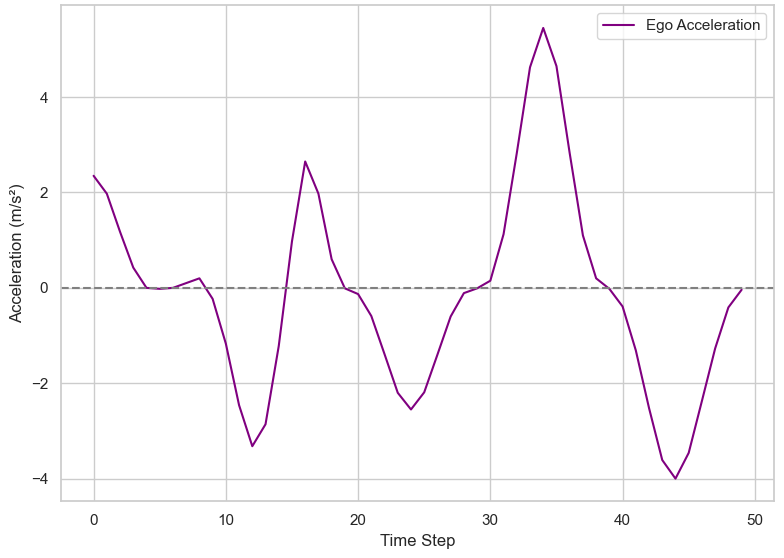}
\caption{Ego Vehicle Acceleration}
\end{subfigure}
\caption{Illustration of Cluster 0: Car-following Instability. (a) Overview of the scene (b) Relative velocity of Front and Ego (c) Acceleration of Ego Vehicle}
\label{fig:cluster0_case}
\end{figure}

This cluster reflects unstable car-following interactions, characterized by close longitudinal spacing and frequent speed modulations by the trailing vehicle.
In Figure~\ref{fig:cluster0_case}(a), two vehicles travel in the same lane with consistently short headway. The relative velocity plot in (b) shows erratic shifts between positive and negative values, indicating that the trailing vehicle alternates between gaining on and falling behind the lead vehicle. These fluctuations reveal a failure to maintain steady pacing, resulting in repeated mismatches in relative speed. The acceleration curve in (c) further confirms this pattern, with dense high-frequency oscillations that suggest continuous and reactive throttle-brake adjustments. Such behavior, though subtle, can reflect poor car-following stability and increase the risk of rear-end conflicts in dense traffic conditions.

\paragraph{\textit{Case Study: Cluster 1.}}

\begin{figure}[htbp]
\centering
\begin{subfigure}[b]{0.15\textwidth}
\includegraphics[height=2.0cm, width=\textwidth]{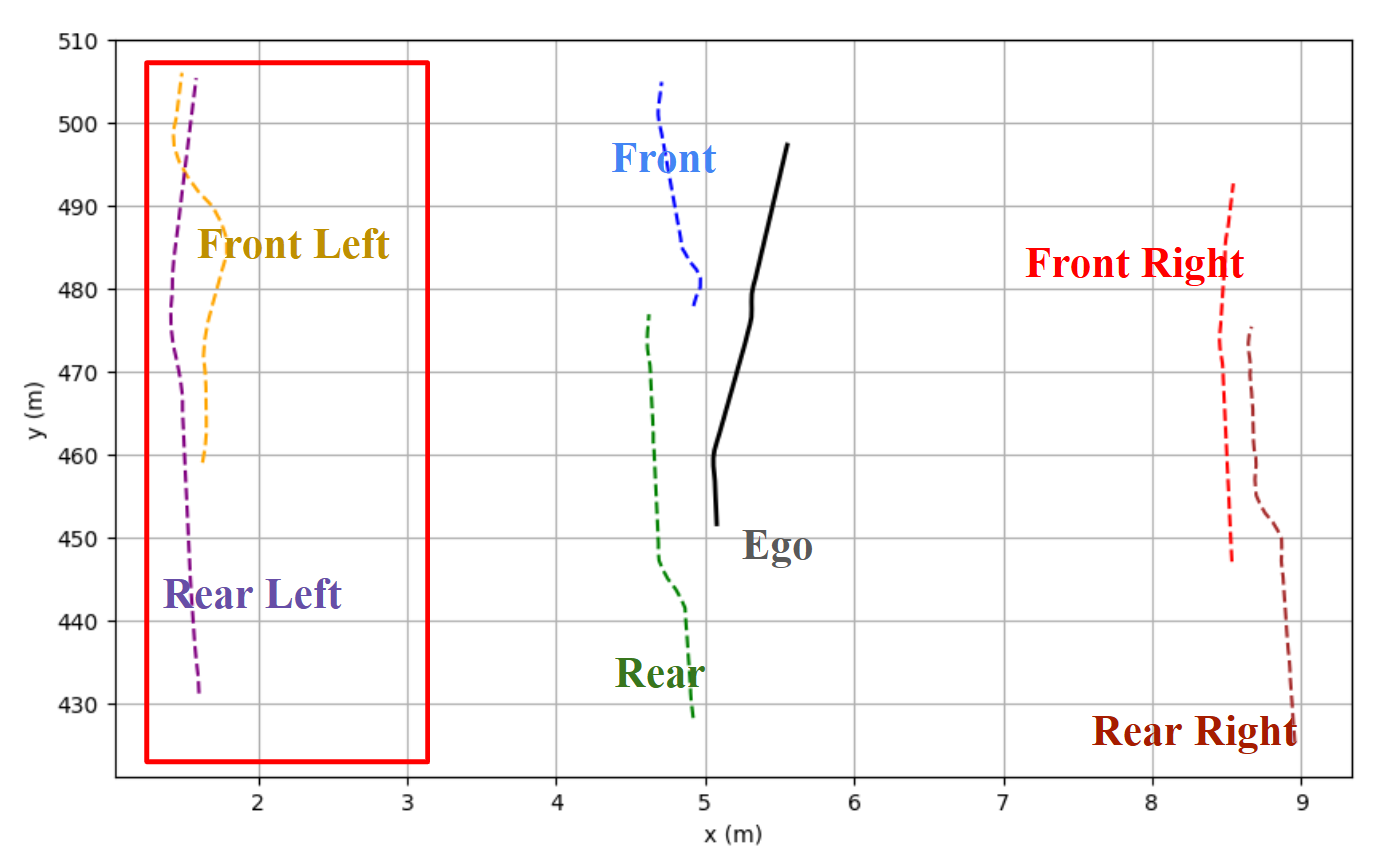}
\caption{Trajectory Snapshot}
\label{fig:cluster1_traj}
\end{subfigure}
\hfill
\begin{subfigure}[b]{0.15\textwidth}
\includegraphics[height=2.0cm, width=\textwidth]{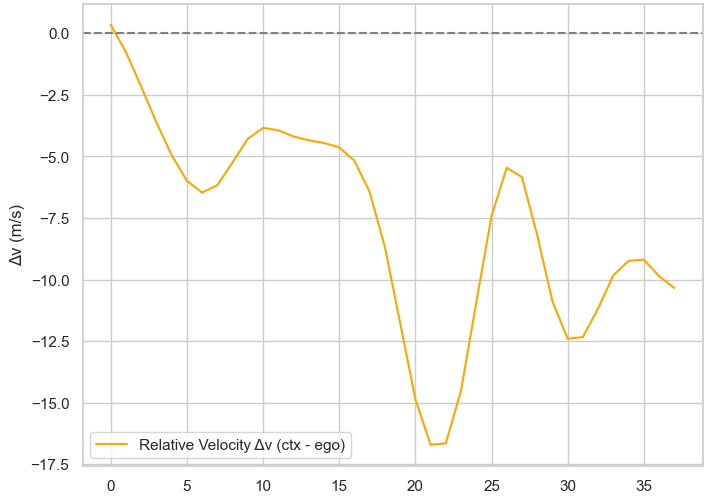}
\caption{Relative Velocity $\Delta v$ }
\label{fig:cluster1_rv}
\end{subfigure}
\hfill
\begin{subfigure}[b]{0.15\textwidth}
\includegraphics[height=2.0cm, width=\textwidth]{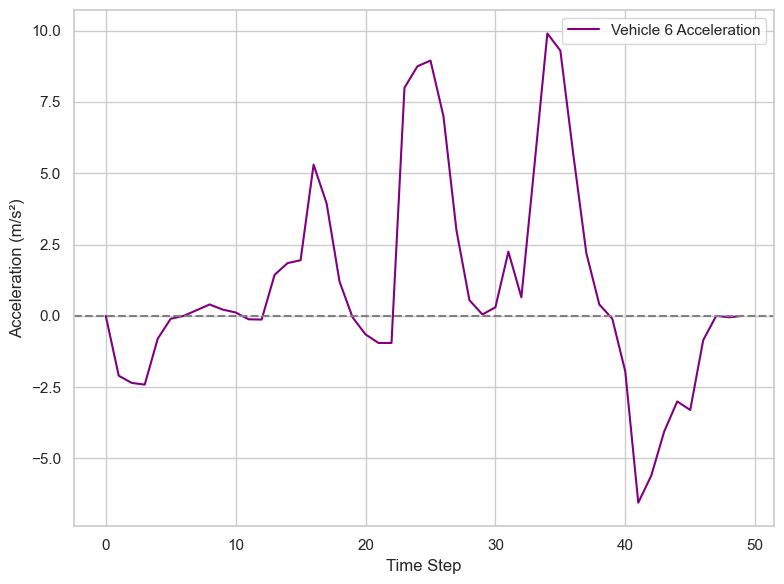}
\caption{Acceleration Curve}
\label{fig:cluster1_acc}
\end{subfigure}
\caption{Illustration of Cluster 1: Sudden Acceleration or Braking. (a) Overview of the scene (b) Relative Velocity of Rear Left and Front Left (c) Acceleration of Rear Left Vehicle}
\label{fig:cluster1_case}
\end{figure}

This case is characterized by sharp longitudinal maneuvers. In Figure~\ref{fig:cluster1_traj}, two vehicles travel in close proximity within the same lane. The acceleration profile in Figure~\ref{fig:cluster1_acc} indicates that the following vehicle exhibits sudden, forceful acceleration episodes. Once close, the vehicle then brakes sharply to avoid collision, creating a distinct acceleration-deceleration cycle. This pattern is further supported by the relative velocity plot in Figure~\ref{fig:cluster1_rv}, which highlights an abrupt increase followed by a quick reversal in speed difference between the two vehicles. Such behavior may challenge the AV’s ability to react promptly to aggressive longitudinal changes.

\paragraph{\textit{Case Study: Cluster 2.}}
\begin{figure}[htbp]
\centering
\begin{subfigure}[b]{0.15\textwidth}
\includegraphics[height=2.0cm, width=\textwidth]{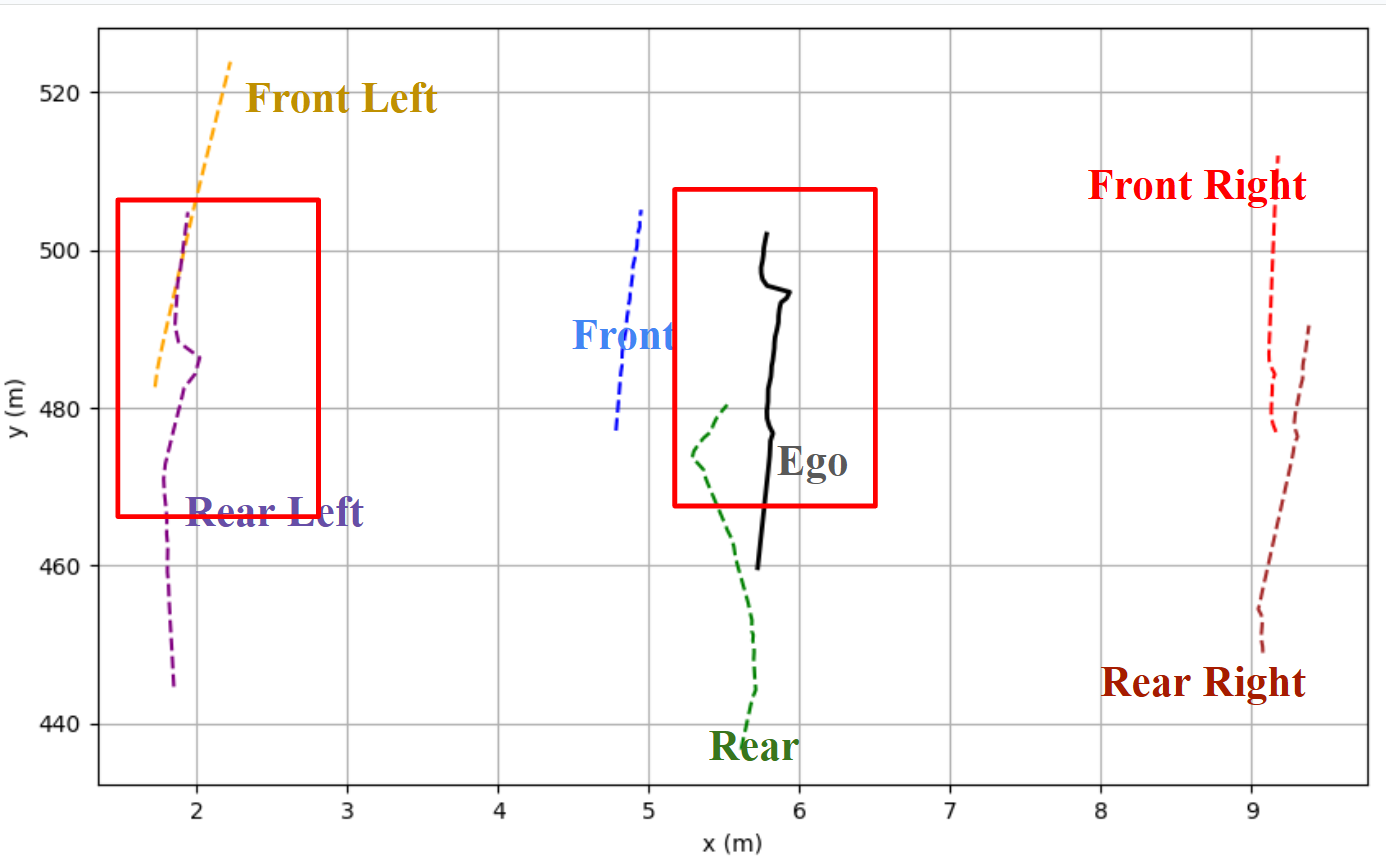}
\caption{Trajectory Snapshot}
\label{fig:cluster2_traj}
\end{subfigure}
\hfill
\begin{subfigure}[b]{0.15\textwidth}
\includegraphics[height=2.0cm, width=\textwidth]{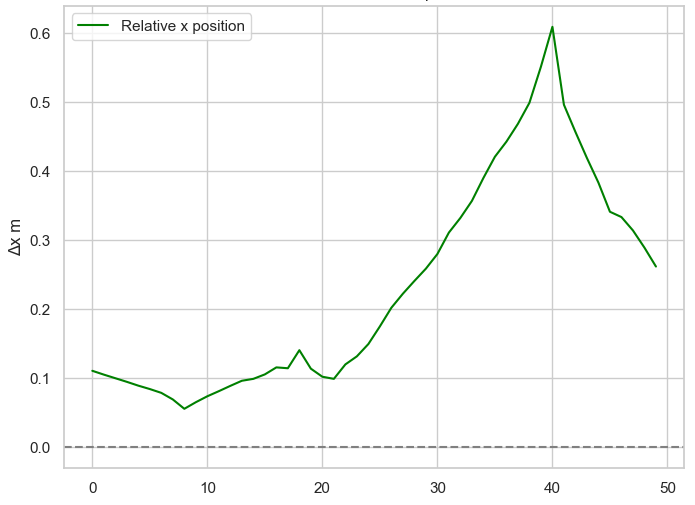}
\caption{Relative Lateral Position $\Delta x$}
\label{fig:cluster2_rx}
\end{subfigure}
\hfill
\begin{subfigure}[b]{0.15\textwidth}
\includegraphics[height=2.0cm, width=\textwidth]{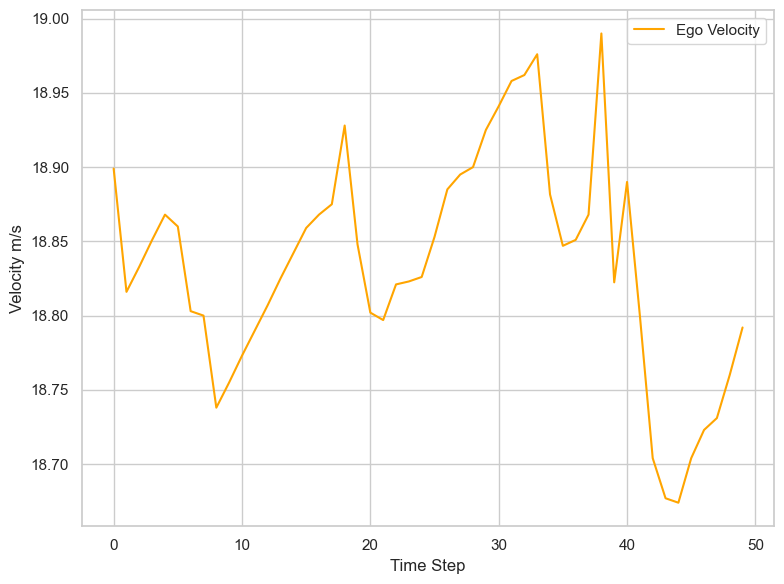}
\caption{Ego Velocity Curve}
\label{fig:cluster2_ev}
\end{subfigure}
\caption{Illustration of Cluster 2: Mild lateral deviation. (a) Overview of the scene (b) Relative Lateral Position of Rear and Ego (c) Velocity of Ego Vehicle}
\label{fig:cluster2_case}
\end{figure}

This pattern illustrates persistent lateral drifting while maintaining stable longitudinal motion. Figure~\ref{fig:cluster2_traj} depicts two neighboring vehicles gradually shifting laterally over time. The $\Delta x$ curve in Figure~\ref{fig:cluster2_rx} confirms this, with a rising and falling profile that peaks around frame 40, suggesting a temporary lane departure or unintentional drift. Meanwhile, the velocity in Figure~\ref{fig:cluster2_ev} remains nearly constant, ruling out braking or evasive causes. Such behaviors may be indicative of inattentiveness or lane-keeping deficiencies.

\paragraph{\textit{Case Study: Cluster 3.}}
\begin{figure}[htbp]
\centering
\begin{subfigure}[b]{0.15\textwidth}
\includegraphics[height=2.0cm, width=\textwidth]{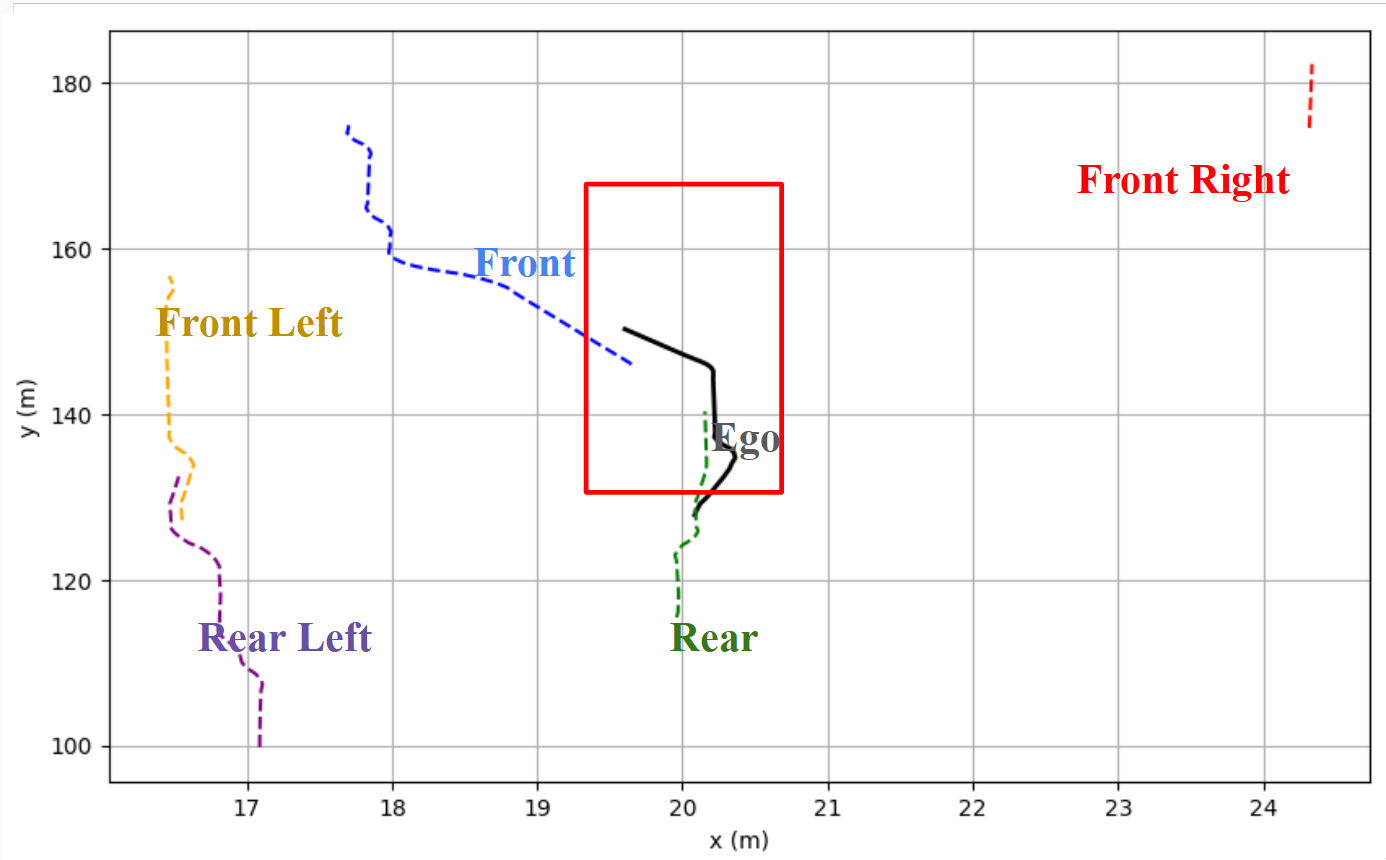}
\caption{Trajectory Snapshot}
\label{fig:cluster3_traj}
\end{subfigure}
\hfill
\begin{subfigure}[b]{0.15\textwidth}
\includegraphics[height=2.0cm, width=\textwidth]{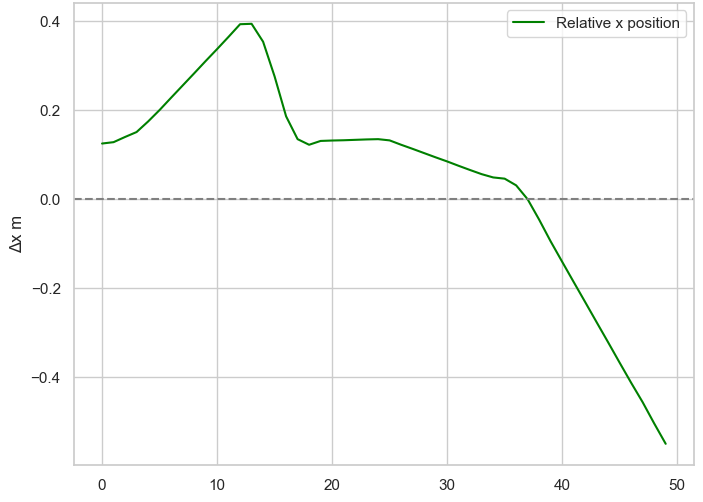}
\caption{Relative Lateral Position $\Delta x$ }
\label{fig:cluster3_rx}
\end{subfigure}
\hfill
\begin{subfigure}[b]{0.15\textwidth}
\includegraphics[height=2.0cm, width=\textwidth]{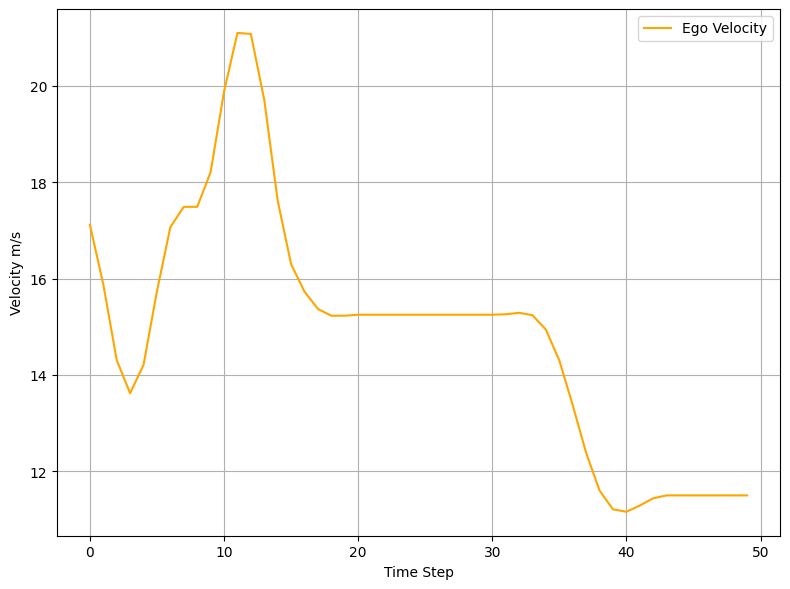}
\caption{Ego Velocity Curve}
\label{fig:cluster3_ev}
\end{subfigure}
\caption{Illustration of Cluster 3: Abrupt lateral maneuver. (a) Overview of the scene (b) Relative Lateral Position of Rear and Ego (c) Velocity of Ego Vehicle}
\label{fig:cluster3_case}
\end{figure}

This case features a sudden lateral maneuver executed by the ego vehicle. In Figure~\ref{fig:cluster3_traj}, the ego and rear vehicle initially travel in the same lane with slight lateral offset. The ego vehicle then performs an abrupt lane change, causing its trajectory to rapidly cross over the rear vehicle’s path. This behavior is evident in the $\Delta x$ profile in Figure~\ref{fig:cluster3_rx}, where the lateral displacement first increases and then drops sharply below zero, indicating a sudden lateral incursion. Meanwhile, the velocity curve in Figure~\ref{fig:cluster3_ev} remains relatively smooth, confirming that the maneuver occurs without significant longitudinal fluctuation. Such sudden lateral movements can compromise lane-level safety margins and introduce unpredictable dynamics into surrounding traffic.

\subsection{Comparison with Baselines}

To assess the effectiveness of our anomaly detection framework, we compare it with two baselines:

\begin{itemize}
    \item Threshold-based: Flags scenes with minimum time-to-collision (TTC) $<$ 1.5\,s, directly indicating imminent collision risks.
    \item Isolation Forest: Trained on four physical features (\textit{min TTC}, \textit{min distance}, \textit{max $\Delta v$}, \textit{max acceleration}) with 15\% contamination.
\end{itemize}

Our method detects 2,832 anomalous scenes, compared to 3,507 (threshold) and 2,835 (Isolation Forest).
Among them, 388 are unique to our model, 219 overlap with threshold only, 512 with Isolation Forest only, and 1,713 with both, showing that while high-risk scenes are shared, our approach captures subtler behaviors like unstable following, lateral drift, and reactive braking.

\begin{figure}[htbp]
    \centering

    % -------- First row --------
    \begin{subfigure}[b]{0.15\textwidth}
        \includegraphics[width=\linewidth]{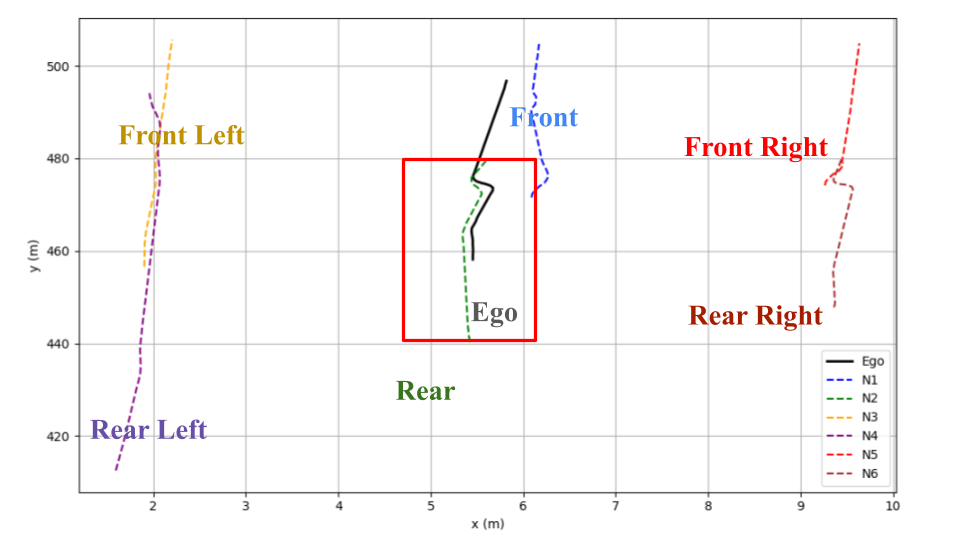}
        \caption{Trajectory snapshot}
    \end{subfigure}
    \hfill
    \begin{subfigure}[b]{0.15\textwidth}
        \includegraphics[width=\linewidth]{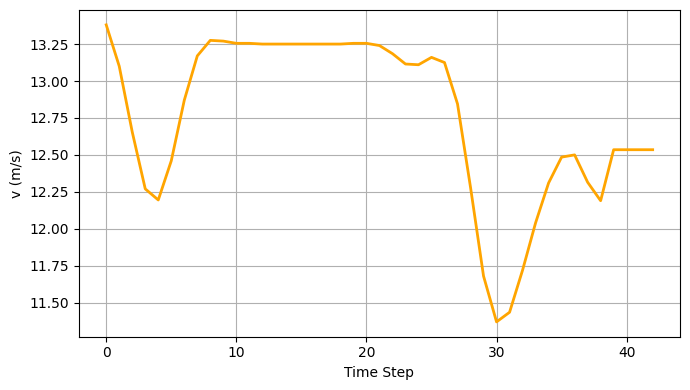}
        \caption{Velocity of front vehicle}
    \end{subfigure}
    \hfill
    \begin{subfigure}[b]{0.15\textwidth}
        \includegraphics[width=\linewidth]{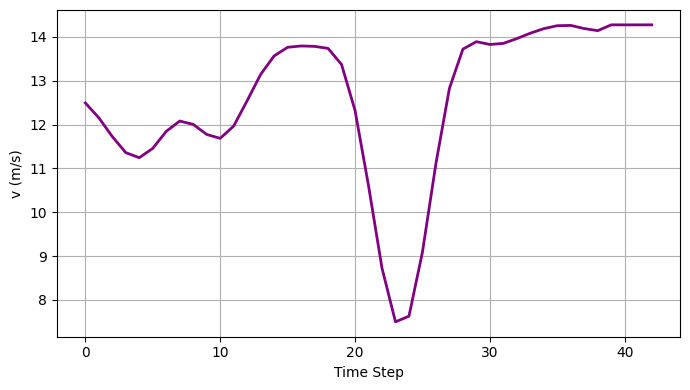}
        \caption{Velocity of ego vehicle}
    \end{subfigure}

    \vspace{0mm} % 可调行距

    % -------- Second row --------
    \begin{subfigure}[b]{0.15\textwidth}
        \includegraphics[width=\linewidth]{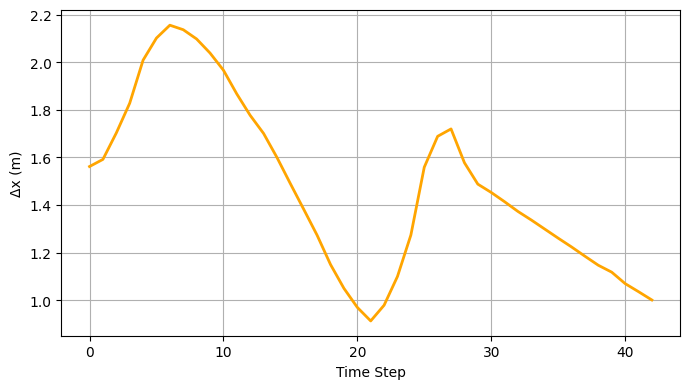}
        \caption{Lateral offset}
    \end{subfigure}
    \hfill
    \begin{subfigure}[b]{0.15\textwidth}
        \includegraphics[width=\linewidth]{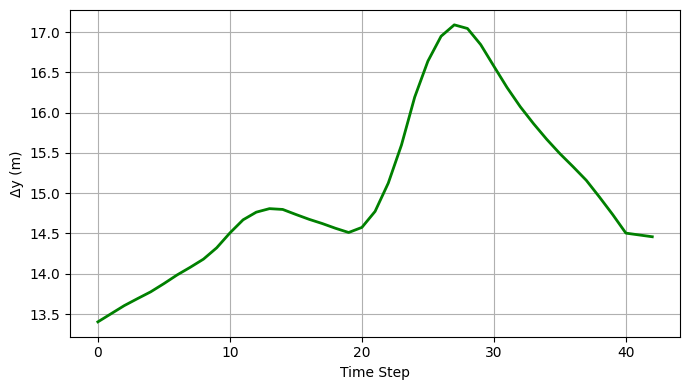}
        \caption{Longitudinal gap }
    \end{subfigure}
    \hfill
    \begin{subfigure}[b]{0.15\textwidth}
        \includegraphics[width=\linewidth]{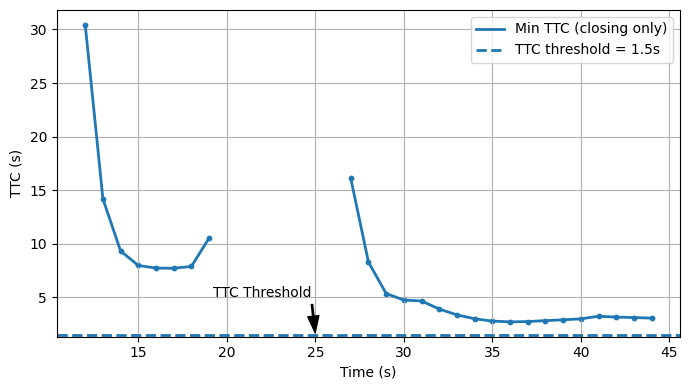}
        \caption{Minimum TTC}
    \end{subfigure}

    \caption{A representative anomaly detected by our method but missed by the TTC-based baseline.}

    \label{fig:missed_case}
\end{figure}

Figure~\ref{fig:missed_case} shows a unique case missed by both baselines: the lead vehicle drifts laterally, prompting abrupt braking from the follower without a TTC violation, highlighting our model’s ability to detect nuanced interaction anomalies.

\section{Discussion and Future Work}

Our results suggest that prediction-residual anomalies capture deviations from learned normal driving patterns that are often consistent with risk-relevant surrogate signatures. The superior performance of the max aggregation suggests that many safety-relevant events manifest as short, high-magnitude deviation spikes rather than sustained average deviations.
The proposed dual evaluation framework ensures both statistical stability and physical interpretability, while K-Means clustering demonstrates how anomaly scores can be transformed into actionable insights for autonomous driving testbeds.

The current evaluation focuses on the NGSIM dataset and an offline setting. 
Future work will extend the framework to more diverse datasets (e.g., Waymo), explore online deployment, employ generative models such as GAIL to enrich safety-critical scenarios, integrate stronger prediction models for sharper residuals, and investigate real-time adaptation for early risk detection.

% \section*{Acknowledgment}

% \section*{References}

\bibliographystyle{IEEEtran}
\bibliography{IEEE-reference}

\end{document}